# A Study for the Feature Core of Dynamic Reduct

Jiayang Wang
(College of Information Science and Engineering, Central South University,
Changsha, Hunan, P.R.China 410083)
csuwjy@mail.csu.edu.cn

Abstract—To the reduct problems of decision system, the paper proposes the notion of dynamic core according to the dynamic reduct model. It describes various formal definitions of dynamic core, and discusses some properties about dynamic core. All of these show that dynamic core possesses the essential characters of the feature core.

Keywords: Rough set, Dynamic reduct, Feature core

## I. INTRODUCTION

Knowledge reduct is an important step in knowledge discovery, and also a favorable method to extract the more generalized rules. There are a number of research papers[1]-[3] about reduct problems, but most of them are just for static reduct.

The reduct methods based on standard rough set are effective to some extent, but there are also some problems to be solved in practice. Standard rough set methods are not always sufficient for extracting laws from decision system. One of the reasons is that these methods are not taking into account the fact that part of the reduct set is chaotic, in other words it is not stable. Dynamic data, incremental data and noise data make the analysis results instable and uncertain. All of these limit the application of rough set theory.

## II. DYNAMIC REDUCT

To the problems of standard rough set reducts or static reducts, dynamic reducts can put up better performance in very large dataset, and also enhances effectively the ability to accommodate noise data.

The rules calculated by means of dynamic reducts are better predisposed to classify unseen cases, because these reducts are in some sense the most stable reducts, and they are the most frequently appearing reducts in sub-decision system created by random samples of a given decision system.

Definition 1: Decision system S=(U, C∪D), where U is a non-empty, finite set called the universal, C is condition attributes set, D is decision attributes set, B=(U', C∪D), U'⊆U, B is called sub-decision system of S. Let $\rho(S)$ denote the set of all sub-decision system of S, F⊆$\rho$(S) is called a F family of decision system S.

Definition 2: Decision system S=(U, C∪D), where U is a non-empty, finite set called the universal, C is condition attributes set, D is decision attributes set, RED(S) denotes the set which contains all reducts of decision system S, and RED(B) denotes the set which includes all reducts of sub-decision system B.

A decision system at least exists one reduct, which is just itself, so the set of reduct is not empty. In many cases, a given decision system may exist several reducts. Each reduct can product a rule set, and it is difficult to justify which is the best rule set. Therefore it is a important to search the most stable reduct, dynamic reduct is proposed in this case.

Definition 3: Decision system S=(U, C∪D), where U is a non-empty, finite set called the universal, C is condition attributes set, D is decision attributes set, F⊆$\rho$(S), dynamic reduct of decision system S is denoted as DR(S, F) [5], and

$$DR(S, F) = RED(S) \cap \bigcap_{B \in F} RED(B).$$

Any element of DR(S, F) is called an F-dynamic reduct, which describes the most stable reducts in decision system. From the definition of dynamic reduct it follows that a relative reduct of S is dynamic if it is also a reduct of all sub-decision system from a given family F by random sampling.

The reducts in a decision system are not stable, sensitive for sample data[4]. Bazan gives the concept and method about dynamic reduct[5], which grounds the most stable reduct of decision system in theory, then the dynamic core is put forward in the paper.

## III. DYNAMIC CORE

Attributes reduct is the basic problem in rough set theory, and the computation of feature core is especially important for resolving this problem. All attributes in the feature core will be presence in any reduct, otherwise discernible relation in decision system can not be ensured. Many reduct algorithms are based on the feature core. According to the feature core one can construct reduct heuristically, and the efficiency of reduct can be improved greatly.

Many references discuss about the feature core of reduct[2][3], but it is just static reduct. For dynamic reduct the feature core still need to be probed in a further step.

Given decision system $S=(U, C \cup D)$, where U is a non-empty, finite set called the universal, C is condition attributes set, D is decision attributes set, we know that the feature core of decision system S in static reduct is

$$CORE(S) = \bigcap_{R \in RED(S)} R \;,$$

the feature core of sub-decision system B is

$$CORE(B) = \bigcap_{R \in RED(B)} R \;.$$

Definition 4: Decision system $S=(U, C \cup D)$, where U is a non-empty, finite set called the universal, C is condition attributes set, D is decision attributes set, $F \subseteq \rho(S)$, the dynamic core of S based on family F is defined by

$$DCORE(S, F) = CORE(S) \cap \bigcap_{B \in F} CORE(B) \;,$$

DCORE(S, F) is called F–dynamic core of decision system S.

Theorem 1: Decision system $S=(U, C \cup D)$, where U is a non-empty, finite set called the universal, C is condition attributes set, D is decision attributes set, $F \subseteq \rho(S)$, then the intersection of F-dynamic reduct contains F-dynamic core, that is

$$\bigcap_{R \in DR(S,F)} R \supseteq DCORE(S, F).$$

Proof: According to the definition of dynamic reduct[4], it follows

$$DR(S, F) = RED(S) \cap \bigcap_{B \in F} RED(B).$$

For any $B \in F$, $DR(S, F) \subseteq RED(B)$, it holds $DR(S, F) \subseteq RED(S)$,

then,

$$\bigcap_{R \in DR(S,F)} R \supseteq \bigcap_{R \in RED(S)} R, \quad \bigcap_{R \in DR(S,F)} R \supseteq \bigcap_{R \in RED(B)} R,$$

$$\bigcap_{R \in DR(S,F)} R \supseteq CORE(S), \quad \bigcap_{R \in DR(S,F)} R \supseteq CORE(B),$$

$$\bigcap_{R \in DR(S,F)} R \supseteq CORE(S) \cap \bigcap_{B \in F} CORE(B),$$

therefore,

$$\bigcap_{R \in DR(S,F)} R \supseteq DCORE(S, F).$$

Theorem 1 means that each attribute in F dynamic core is included by all F dynamic reducts. Dynamic reducts is the most stable reducts of decision system S, then dynamic core is the most stable core of decision system S, which represents a stable set of unreduced attributes.

## IV. (F-λ)–DYNAMIC CORE

F–dynamic core can be sometimes too much restrictive so here applies a generalization of F–dynamic core. It will be more suitable for noise data.

Definition 5: Decision system $S=(U, C \cup D)$, where U is a non-empty, finite set called the universal, C is condition attributes set, D is decision attributes set, $F \subseteq \rho(S)$, $\lambda \in (0.5, 1]$, the (F-$\lambda$)–dynamic core of decision system S based on family F is defined by

$$DCORE_\lambda(S, F) = \{a \in CORE(S) \mid \frac{|\{B \in F : a \in CORE(B)\}|}{|F|} \geq \lambda\}.$$

$\lambda$ is precision coefficient, and the value of $\lambda$ decides which attribute belongs to (F-$\lambda$)–dynamic core $DCORE_\lambda(S, F)$.

$\lambda$ approaches 1, $DCORE_\lambda(S, F)$ will be closed to DCORE(S, F), while $\lambda$ approaches 0.5, $DCORE_\lambda(S, F)$ is more rough compared with CORE(S, F), but $DCORE_\lambda(S, F)$ will comprise more attributes.

Theorem 2: Decision system $S=(U, C \cup D)$, where U is a non-empty, finite set called the universal, C is condition attributes set, D is decision attributes set, F、F'$\subseteq \rho(S)$, we have the following propositions.

(1) If F={S}, then DCORE(S, F)=CORE(S)

In this case dynamic core is just the feature core of decision system S.

(2) $DCORE_1(S, F) = DCORE(S, F)$

Actually, when $\lambda$ increases from 0.5 to 1, the dynamic core will change from $DCORE_\lambda(S, F)$ to DCORE(S, F).

(3) If $\lambda \leq \lambda_1$, then $DCORE_{\lambda_1}(S, F) \subseteq DCORE_\lambda(S, F)$

Obviously, for any $\lambda \in (0.5, 1]$, there will be DCORE(S, F)$\subseteq DCORE_\lambda(S, F)$

Theorem 3: Decision system $S=(U, C \cup D)$, where U is a non-empty, finite set called the universal, C is condition attributes set, D is decision attributes set, $F \subseteq \rho(S)$, $\lambda \in (0.5, 1]$, then the intersection of (F-$\lambda$)–dynamic reduct contains (F-$\lambda$)–dynamic core.

Proof: According to the definition of (F-$\lambda$)–dynamic core, it follows

$$CORE(S) \supseteq DCORE_\lambda(S, F).$$

(F-$\lambda$)–dynamic reduct is

$$DR_\lambda(S, F)=\{Q\in RED(S) \mid \frac{|\{B\in F: Q\in RED(B)\}|}{|F|}\geq\lambda\}^{[4]}.$$

It is obviously that $DR_\lambda(S, F)\subseteq RED(S)$,

then, $\bigcap_{R\in DR_\lambda(S,F)} R \supseteq \bigcap_{R\in RED(S)} R = CORE(S)$,

therefore, $\bigcap_{R\in DR_\lambda(S,F)} R \supseteq DCORE_\lambda(S, F)$.

## V. GENERALIZED DYNAMIC CORE

According to the definition of dynamic core, if some feature attributes of any sub-decision system in F family are comprised by dynamic core, then it is certainly a feature attribute of decision system S. This notion can be sometimes not convenient because we are interested in useful sets of attributes which are not necessarily reducts of the decision system. Therefore we have to generalize the notion of a dynamic core.

Definition 6: Decision system $S=(U, C\cup D)$, where U is a non-empty, finite set called the universal, C is condition attributes set, D is decision attributes set, $F\subseteq\rho(S)$, then

$$GDCORE(S, F)= \bigcap_{B\in F} CORE(B)$$

GDCORE(S, F) is called the generalized dynamic core of decision system S.

Definition 7: Decision system $S=(U, C\cup D)$, where U is a non-empty, finite set called the universal, C is condition attributes set, D is decision attributes set, $F\subseteq\rho(S)$, $\lambda\in(0.5, 1]$, then

$$GDCORE_\lambda(S, F)=\{a\in C \mid \frac{|\{B\in F: a\in CORE(B)\}|}{|F|}\geq\lambda\}$$

The $GDCORE_\lambda(S, F)$ is called the (F-$\lambda$)–generalized dynamic core of decision system S.

Theorem 4: Decision system $S=(U, C\cup D)$, where U is a non-empty, finite set called the universal, C is condition attributes set, D is decision attributes set, $F \subseteq \rho(S)$, $\lambda \in (0.5, 1]$, we have the following propositions.

(1) $DCORE(S, F) \subseteq GDCORE(S, F)$

By definition 6 and 7 we know that it is obviously true.

(2) $DCORE_\lambda(S, F) \subseteq GDCORE_\lambda(S, F)$

By (1) it is also obviously true.

(3) If $S \in F$, then $GDCORE(S, F) = DCORE(S, F)$

When F family contains decision system S, then generalized dynamic core will be just dynamic core.

Theorem 5: Decision system $S=(U, C \cup D)$, where U is a non-empty, finite set called the universal, C is condition attributes set, D is decision attributes set, $F \subseteq \rho(S)$, $\lambda \in (0.5, 1]$, then

(1) The intersection of F–generalized dynamic reduct contains F–generalized dynamic core

Proof: According to the definition of F–dynamic reduct[4], it follows

$$GDR(S, F) = \bigcap_{B \in F} RED(B),$$

for any $B \in F$, $GDR(S, F) \subseteq RED(B)$,

then, $\bigcap_{R \in GDR(S,F)} R \supseteq \bigcap_{R \in RED(B)} R$,

$$\bigcap_{R \in GDR(S,F)} R \supseteq CORE(B) \supseteq \bigcap_{B \in F} CORE(B),$$

therefore,

$$\bigcap_{R \in GDR(S,F)} R \supseteq GDCORE(S, F).$$

(2) The intersection of (F-λ)–generalized dynamic reduct contains (F-λ)–generalized dynamic core

Proof: For any attribute $a \in GDCORE_\lambda(S, F)$, it satisfies $\dfrac{|\{B \in F : a \in CORE(B)\}|}{|F|} \geq \lambda$,

It might as well suppose that attribute a exists in core of k sub-decision systems,

$$k = |\{B \in F: a \in CORE(B)\}|,$$

we have

$$a \in CORE(B_1), a \in CORE(B_2), \cdots\cdots, a \in CORE(B_k),$$

$$a \in \bigcap_{R \in RED(B_1)} R, \ a \in \bigcap_{R \in RED(B_2)} R, \cdots\cdots, a \in \bigcap_{R \in RED(B_k)} R \qquad (1)$$

According to the definition of (F-λ)–generalized dynamic reduct[4],

$$GDR_\lambda(S, F) = \{R \subseteq S \mid \frac{|\{B \in F: R \in RED(B)\}|}{|F|} \geq \lambda\},$$

because $\lambda > 0.5$, any reduct R in (F-λ)–generalize dynamic reduct satisfies:

$$R \in RED(B_1) \cup RED(B_2) \cup \cdots \cup RED(B_k).$$

By (1), for any attribute a, we can infer: $a \in \bigcap_{R \in GDR_\lambda(S,F)} R$,

then, $\bigcap_{R \in GDR_\lambda(S,F)} R \supseteq GDCORE_\lambda(S, F)$.

Actually, supposes $a \notin \bigcap_{R \in GDR_\lambda(S,F)} R$,

there exists $R' \in GDR_\lambda(S, F)$, and $a \notin R'$.

By (1), it follows

$$R' \notin RED(B_1), R' \notin RED(B_2), \cdots\cdots, R' \notin RED(B_k) \qquad (2)$$

If $\lambda > 0.5$, then (2) is contradiction with hypothesis,

Therefore, it must be $a \in \bigcap_{R \in GDR_\lambda(S,F)} R$.

Only while $\lambda \leq 0.5$, it may be true for the hypothesis, but we have fixed $\lambda > 0.5$, so it is impossible.

## VI. CONCLUSION

The notion of dynamic core is presented based on dynamic reducts in the paper. Dynamic core consists with the most stable attributes which can not be reduced, and describes the set of feature attributes. At the same time it is proved that the intersection of dynamic reducts contains dynamic core, which is suitable for all kinds of definitions about dynamic reducts. All of these show that the dynamic core possesses the essential properties of the feature core in deed. We can say that dynamic core expresses the feature attributes in a more general way.